\documentclass[a4paper,twoside]{article}

\usepackage[FIGTOPCAP,normal]{subfigure} 
\usepackage{url}
\usepackage[utf8]{inputenc}
\usepackage{epsfig}
\usepackage{calc}
\usepackage{amssymb}
\usepackage{amstext}
\usepackage{amsmath}
\usepackage{amsthm}
\usepackage{multicol}
\usepackage{pslatex}
\usepackage{apalike}
\usepackage{SCITEPRESS}
\usepackage[small]{caption}

\usepackage[english]{babel} 
\usepackage{tikz}
\usepackage{graphicx}
\usetikzlibrary{calc,shapes.multipart,chains,arrows,fit,shapes,calc,backgrounds,trees}

\begin{document}

\title{Combining Simulated Annealing and Monte Carlo Tree Search for Expression Simplification}

\author{\authorname{Ben Ruijl\sup{1,2}, Jos Vermaseren\sup{2}, Aske Plaat\sup{1} and Jaap van den Herik\sup{1}}
\affiliation{\sup{1}Tilburg University, Tilburg center for Cognition and Communication, Warandelaan 2, 5037 AB Tilburg, The Netherlands}
\affiliation{\sup{2}Nikhef Theory Group, Science Park 105, 1098 XG Amsterdam, The Netherlands}
}


\abstract{In many applications of computer algebra large expressions must be simplified to make repeated numerical evaluations tractable. Previous works presented heuristically guided improvements, e.g., for Horner schemes. The remaining expression is then further reduced by common subexpression elimination. A recent approach successfully applied a relatively new algorithm, Monte Carlo Tree Search (MCTS) with UCT as the selection criterion, to find better variable orderings. Yet, this approach is fit for further improvements since it is sensitive to the so-called ``exploration-exploitation'' constant $C_p$ and the number of tree updates $N$. In this paper we propose a new selection criterion called Simulated Annealing UCT (SA-UCT) that has a \emph{dynamic} exploration-exploitation parameter, which decreases with the iteration number $i$ and thus reduces the importance of exploration over time. First, we provide an intuitive explanation in terms of the exploration-exploitation behavior of the algorithm. Then, we test our algorithm on three large expressions of different origins. We observe that SA-UCT widens the interval of good initial values $C_p$ where best results are achieved. The improvement is large (more than a tenfold) and facilitates the selection of an appropriate $C_p$. 
}

\onecolumn \maketitle \normalsize \vfill

\section{\uppercase{Introduction}}
\label{sec:introduction}

\noindent In High Energy Physics (HEP) expressions with millions of terms arise from the calculation of processes described by Feynman diagrams. Typically, these expressions have to be numerically integrated to predict cross sections and particle decays in collision processes. For example, in the Large Hadron Collider in CERN such calculations were essential to confirm the likely existence of the Higgs boson.

In order to predict the effects of currently undiscovered particles and to improve the accuracy of current HEP models, higher-order loop corrections are needed, causing the size of the expressions to grow exponentially \cite{Peskin1995}. The intermediate forms of these expressions may often take terabytes of disk space. Novel approaches are required to simplify these expressions to make evaluation feasible.

To simplify expressions Horner schemes and common subexpression elimination (CSEE) may be used. Horner's rule for simplifying expressions goes back to 1819 \cite{horner1819}. CSEE is commonly used in compiler construction \cite{dragon}. In \cite{Kuipers2013} the first application of MCTS for finding a better variable ordering was presented, using UCT~\cite{Kocsis2006} as the selection criterion (see section \ref{sec:sa-uct}). The MCTS performance is sensitive to the choice of three parameters: $C_p$, $N$, and $R$. $C_p$ is the constant that governs the exploration-exploitation choices of the algorithm, $N$ is the number of tree updates, and $R$ is the number of times the MCTS is repeated. At the previous ICAART conference the sensitivity to $C_p$ and $N$ was presented \cite{Herik2013}. At CCIS/BNAIC the sensitivity to $R$ was recognized \cite{Herik2013B,Kuipers2013B}. We believe that the practical applicability of the algorithm will improve as the sensitivity to these parameters is harmonized. 

This paper focuses on $C_p$. We modify the UCT formula by introducing an exploration-exploitation parameter $T(i)$, which decreases with the current iteration number $i$, effectively making the constant $C_p$ a variable $T(i)$. As a result, the first iterations will be explorative and throughout the search, the child selection will gradually become more exploitative, favoring optimizing a local minimum over exploration. The parameter $T$ is similar to the role of the temperature in simulated annealing (hence the name $T$). We refer to the new formula as Simulated Annealing UCT (SA-UCT).

We have tested our algorithms on three large expressions of different origins and we observed that SA-UCT widens the interval of good initial temperatures $T(0)$, where the number of operations is near the global minimum, by more than a tenfold for all three test expressions.

The paper is structured as follows. Section \ref{sec:background} provides a background and related work on expression simplification and MCTS. Section \ref{sec:sa-uct} presents our new selection criterion called SA-UCT. Section \ref{sec:results} shows our measurement results. Section \ref{sec:conclusion} presents the conclusion and section \ref{sec:futurework} gives an outlook on future work.

\section{\uppercase{Background}}
\label{sec:background}
\noindent Numerous methods for simplifying expressions have been proposed. Here we mention Horner schemes \cite{Knuth1997}, common subexpression elimination \cite{dragon}, Breuer's growth algorithm for systems of expressions \cite{Breuer1969}, and partial syntactic factorization \cite{Leiserson2010}. In this paper we focus on two of these: Horner schemes and common subexpression elimination.

\subsection{Horner schemes}
One elementary method of reducing the number of multiplications in an expression is based on Horner's rule \cite{horner1819,Knuth1997,Ceberio2004}. Horner's rule is straightforwardly lifting a variable outside brackets. For multivariate expressions Horner's rule can be applied multiple times, once for each variable. The order of the extracted variables is called a Horner scheme. For example:
\begin{equation}
x^3y^2 + x^2y + x^3z \Rightarrow x^2(y + x(y^2 + z))
\end{equation}
By twice extracting the variable $x$ (i.e., $x^2$ and $x$), the number of multiplications is reduced from $9$ to $4$. The number of additions remains the same, which is a general property of Horner schemes.

In multivariate expressions with $n$ variables, there are $n!$ ways of extracting variables. For example, the above expression could also be transformed to 
\begin{equation}
x^3z + y(x^2(1 + xy))
\label{eq:horner}
\end{equation}
by first extracting $y$ and then $x$. Using this scheme, we have $7$ multiplications left. Thus, this Horner scheme is inferior to the first one. 

The problem of selecting an optimal ordering is NP-hard \cite{Ceberio2004}. A heuristic that works reasonably well is to select the variables according to their frequency of occurrence (``occurrence order''), see e.g., \cite{Kuipers2013}. However, this does not always yield good results, also not when combined with common subexpression elimination (see below). 

\subsection{Common subexpression elimination}
\label{ch:csee}
\begin{figure}
\centering
\includegraphics{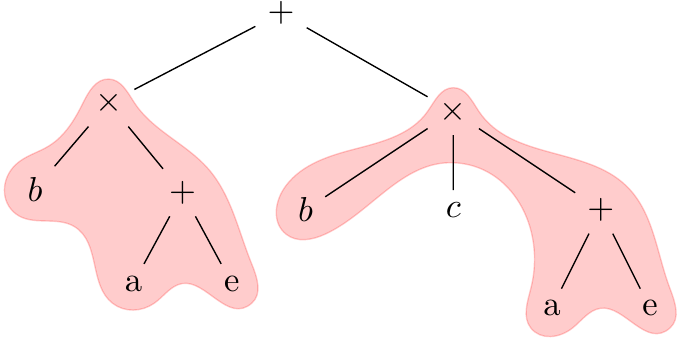}
\caption{A common subexpression (shaded) in a commutative and associative tree.}
\label{cse_tree}
\end{figure}

A way to reduce the number of operations even further is to perform a common subexpression elimination (CSEE). This strategy is well known in the field of compiler construction \cite{dragon}. CSEE creates new symbols for each subexpression that appears twice or more. Consequently, the subexpression has to be computed only once. Figure \ref{cse_tree} shows an example of a subexpression in a tree representation.

We note that there is an interplay between Horner and CSEE in the following example:
\begin{align*}
&\sin(x) + \cos(x) + \sin(x)x + \cos(x)x =\\
&\sin(x) + \cos(x) + x(\sin(x) + \cos(x)) =\\
&T + xT
\end{align*}
Most practical methods of detecting common subexpressions will not find $\sin(x) + \cos(x)$ as a subexpression in the first line, whereas in the second line it is detected. Hence, we observe that Horner schemes can expose common subexpressions.

\subsection{Monte Carlo Tree Search}
Because there is an interplay between Horner schemes and CSEE, a trade-off exists between (1) selecting the optimal Horner scheme that reduces the largest number of multiplications and (2) selecting the Horner scheme that exposes the maximum number of CSEs. The contrast is between (1) a Horner scheme that reduces many multiplications, but has few CSEs, and (2) an average Horner scheme that exposes many CSEs. Category (2) would probably reduce the number of operations more than category (1). To find the best option, an optimization method is needed.

Our goal is to minimize the total number of operations after both the Horner scheme and the CSEE have been applied to the expression. Motivated by the successes in \cite{Kuipers2013}, we apply Monte Carlo Tree Search (MCTS) to our set of large expressions. A rich literature exists on MCTS, which is successfully applied in the game of Go \cite{Coulom2006}. For an overview, see, e.g., \cite{Browne2012}.

\tikzset{
  treenode/.style = {align=center, inner sep=0pt, text centered,
    font=\sffamily},
  arn_r/.style = {treenode, circle, black, draw=black, text width=1.5em,semithick,-}, 
  ar/.style = {arn_r,->,>=stealth,very thick},
  arr/.style = {arn_r,<-,>=stealth,very thick}
}

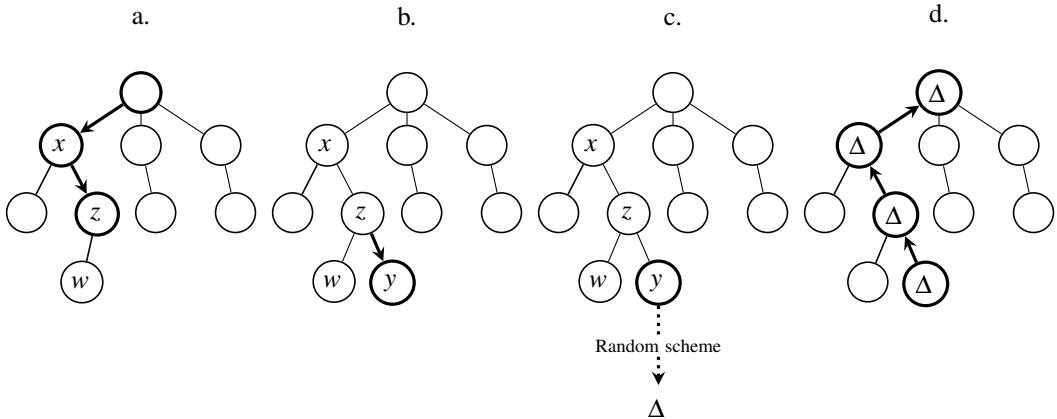
\begin{figure*}[hbt!]
\centering
\begin{tikzpicture}[level/.style={sibling distance = 1.5cm/#1,
  level distance = 1.0cm}] 
  
\begin{scope}[scale=0.7]
\node [ar] (r1) {}
    child[ar]{ node [ar] {$x$} 
            child[arn_r,below left]{ node [arn_r] {}
            }
            child[ar,below right]{ node [ar] {$z$}
							child[arn_r,below left]{ node [arn_r] {$w$}}
            }                            
    }
    child{ node [arn_r] {} 
    	child[below right] { node [arn_r] {} 
    	} 
    }
    child{ node [arn_r] {}
            child[below right]{ node [arn_r] {}
            }
		}
;
\node[above= 0.5cm of r1] {a.};
\end{scope}

\begin{scope}[scale=0.7,shift={(5,0)}]
\node [arn_r] (r2) {}
    child[]{ node [arn_r] {$x$} 
            child[arn_r,below left]{ node [arn_r] {} 
            }
            child[below right]{ node [arn_r] {$z$}
							child[below left]{ node [arn_r] {$w$}}
							child[ar,below right]{ node [ar] {$y$} }
            }                            
    }
    child{ node [arn_r] {} 
    	child[below right] { node [arn_r] {} 
    	} 
    }
    child{ node [arn_r] {}
            child[below right]{ node [arn_r] {}
            }
		}
;
\node[above= 0.5cm of r2] {b.};
\end{scope}

\begin{scope}[scale=0.7,shift={(10,0)}]
\node [arn_r] (r3) {}
    child[]{ node [arn_r] {$x$} 
            child[arn_r,below left]{ node [arn_r] {} 
            }
            child[below right]{ node [arn_r] {$z$}
							child[below left]{ node [arn_r] {$w$}}
							child[below right]{ node[ar] (ex) {$y$} }
            }                            
    }
    child{ node [arn_r] {} 
    	child[below right] { node [arn_r] {} 
    	} 
    }
    child{ node [arn_r] {}
            child[below right]{ node [arn_r] {}
            }
		}
; 
\node[above= 0.5cm of r3] {c.};
\draw[dotted,ar] (ex.south) to node[rectangle,anchor=center, text width=2cm,midway,font=\scriptsize,fill=white] {\text{Random scheme}} ++(0,-1.5cm) node[below] {$\Delta$};
\end{scope}

\begin{scope}[scale=0.7,shift={(15,0)}]
\node [arr] (r4) {$\Delta$}
    child[arr]{ node [arr] {$\Delta$} 
            child[arn_r,below left]{ node [arn_r] {} 
            }
            child[arr,below right]{ node [arr] {$\Delta$}
							child[arn_r,below left]{ node [arn_r] {}}
							child[arr,below right]{ node[arr] (ex) {$\Delta$} }
            }                            
    }
    child{ node [arn_r] {} 
    	child[below right] { node [arn_r] {} 
    	} 
    }
    child{ node [arn_r] {}
            child[below right]{ node [arn_r] {}
            }
		}
; 
\node[above= 0.5cm of r4] {d.};
\end{scope}

\end{tikzpicture}
\caption{An overview of the four phases of MCTS: selection (a), expansion (b), simulation (c), and backpropagation (d). The selection of a not fully expanded node is done using the best child criterion. $\Delta$ is the number of operations left in the final expression, after the Horner scheme and CSEE have been applied. See also \cite{Browne2012}.}
\label{fig:mcts}
\end{figure*}

An outline of the MCTS algorithm is displayed in figure \ref{fig:mcts}. A tree is built in which each node is a variable that will be extracted. The tree will be built iteratively. At each iteration, a leaf (or a not fully expanded node) is chosen according to a selection criterion (see \ref{fig:mcts}(a)) and section \ref{sec:sa-uct}). This node is (further) expanded by randomly picking one of the unvisited children (see \ref{fig:mcts}(b)). Starting from this new leaf, we continue the path by randomly selecting children (i.e., variables) that have not been selected so far (see \ref{fig:mcts}(c)). The \emph{complete} path is our Horner scheme\footnote{Note the difference with games such as Go, where only the first move is needed.}. For this scheme, we compute a score $\Delta$, which is the number of operations after the Horner scheme and CSEE have been applied (see \ref{fig:mcts}(c) again). Finally, the result is propagated backwards through the tree (see \ref{fig:mcts}(d)). For a more detailed explanation of MCTS, see \cite{Browne2012}.

Thus, MCTS is able to capture the trade-off of the Horner scheme and CSEE by using a score $\Delta$ which is the number of operations of the final expression after Horner and CSEE have been applied.

Since we are interested in the best Horner scheme, we keep track of the best path that we come across during the tree updates. This path may not be completed in the tree if the tree did not reach the bottom or if there was another random playout that is better than the (partial) path through the tree.

The essence of MCTS is finding a proper trade-off between exploiting nodes that have been characterized as good and exploring other (new) nodes that may contain a promising path. The challenge of a good algorithm is in balancing the exploration-exploitation issue.

In the next section we modify the exploration part of the UCT selection criterion to scale with the iteration number. In related work a different strategy has been applied to make the importance of exploration versus exploitation iteration-number dependent. For example, Discounted UCB \cite{Kocsis2006B} and Accelerated UCT \cite{Hashimoto2012} both modify the average score of a node (see below) to discount old wins over new ones. In contrast, this work focuses on the exploration-exploitation constant $C_p$.

\section{\uppercase{Our algorithm: SA-UCT}}
\label{sec:sa-uct}
\noindent In many MCTS implementations UCT (eq. \eqref{eq:uct}) is chosen as the selection criterion \cite{Browne2012,Kocsis2006}:
\begin{equation}
\underset{\text{children $c$ of $s$}}{\operatorname{argmax}} \bar{x}(c) + 2 C_p \sqrt{\frac{2 \ln n(s)}{n(c)}}
\label{eq:uct}
\end{equation}
where $c$ is a child node of node $s$, $\bar{x}(c)$ the average score of node $c$, $n(c)$ the number of times the node $c$ has been visited, and $C_p$ the exploration-exploitation constant. This constant determines the probability that a child is selected that does not have a good average score\footnote{In our application, the average score is the number of operations without optimizations divided by the average number of operations for visited paths through this node, see \cite{Kuipers2013}.}, but has not been visited often. If this constant is high, more iterations will be spent on exploration and if this constant is low, the iterations will be spent on exploitation. Generally, a higher $C_p$ results in broader trees, whereas a smaller $C_p$ yields deeper trees.

For our application, it matters that the tree is expanded as deeply as possible, since we want to optimize the \emph{entire} Horner scheme, instead of just selecting the optimal first node, as is the case in games such as Go (please note, this is an important difference). Therefore, it is of higher value that the last iterations are used to deepen the tree and improve the current local minimum than performing additional explorations. To achieve this, we introduce a \emph{dynamic} exploration-exploitation parameter $T$ (for Temperature) that linearly decreases with the number of iterations:

\begin{equation}
T(i) =  C_p \frac{N - i}{N}
\label{eq:t}
\end{equation}
where $i$ is the current iteration number, $N$ the preset maximum number of iterations, and $C_p$ the initial exploration-exploitation constant at $i=0$. 

Our new best child criterion becomes:
\begin{equation}
\underset{\text{children $c$ of $s$}}{\operatorname{argmax}} \bar{x}(c) + 2 T \sqrt{\frac{2 \ln n(s)}{n(c)}}
\label{eq:sa-uct}
\end{equation}
where $c$ is a child of node $s$, $\bar{x}(c)$ is the average score of child $c$, $n(c)$ the number of visits at node $c$, and $T$ the dynamic exploration-exploitation parameter of eq.~(\ref{eq:t}).

Thus, the first iterations are used for exploration and gradually the focus shifts to exploitation and optimization of the currently found local minimum. This process can be thought of as a variant of simulated annealing \cite{Kirkpatrick1983}, where a temperature determines the probability of exploring energetically unfavorable states. Starting at high temperatures, there is a great deal of exploration and when the temperature gradually decreases, the system converges to a local minimum. In our case, the decreasing exploration-exploration parameter $T$ takes the role of the temperature.
Because of the similarity between these approaches, we call eq.~\eqref{eq:sa-uct} ``Simulated Annealing UCT (SA-UCT)''.

\section{\uppercase{Results}}
\label{sec:results}
\noindent In \cite{Kuipers2013}, a sensitivity analysis of different parameters of MCTS is presented, which shows that there is a small interval of $C_p$ for which the number of operations is close to the global best. We call this the \emph{region of interest}. Below we investigate how this region changes if we use SA-UCT as selection criterion. Our experimental setup is as follows: we compare the number of operations after the Horner scheme and CSEE have been applied for fixed $N$ and different $C_p$ for SA-UCT (eq.~\eqref{eq:sa-uct}) with those for the original UCT (eq.~\eqref{eq:uct}). In SA-UCT, $C_p$ is the starting value (initial temperature) $T(0)$. We randomly sample 4000 dots for each graph (not to be confused with the number of operations on the y-axis that also starts with 4000).   

We shall perform a sensitivity analysis of $C_p$ on the number of operations for three expressions from mathematics and physics, namely HEP($\sigma$), res(7,5), and F13, see \cite{Kuipers2013}. HEP($\sigma$) and F13 arise from parts of different Feynman diagrams and res(7,5) is a resultant (an object commonly used in number theory). 

\begin{figure*}
\centering
\textbf{HEP($\sigma$) with 15 variables} \par\medskip
\subfigure[$N=300$]{\includegraphics[scale=0.62]{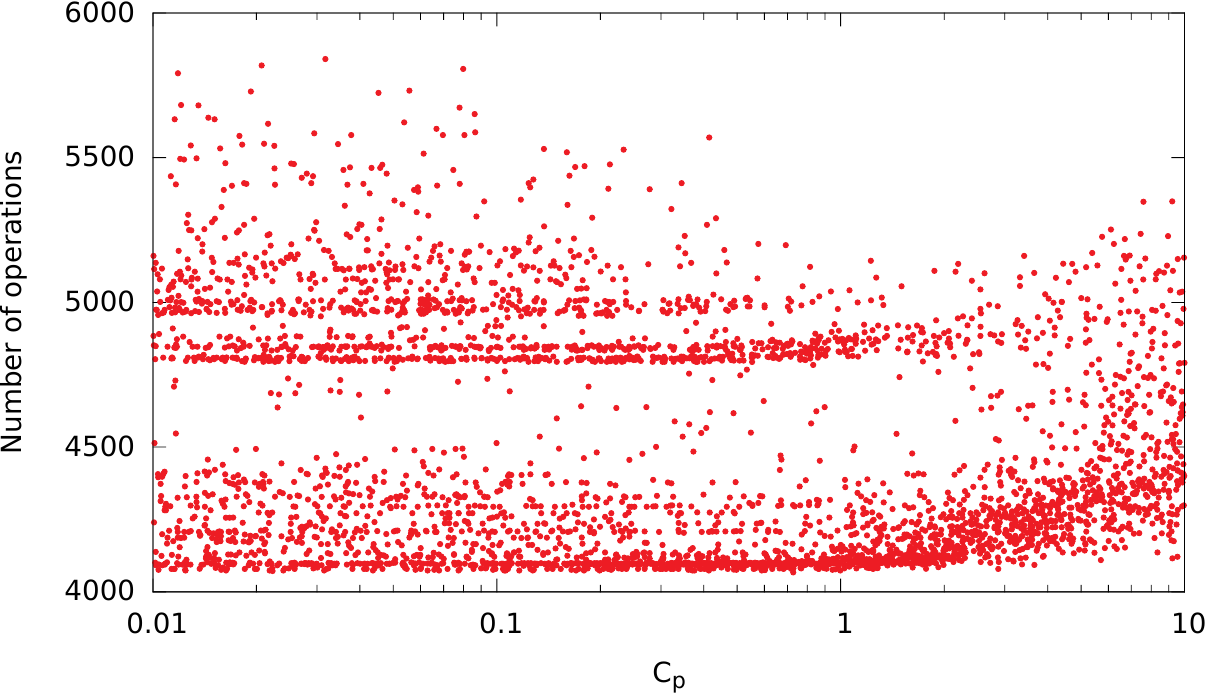}}
\subfigure[$N=300$]{ \includegraphics[scale=0.62]{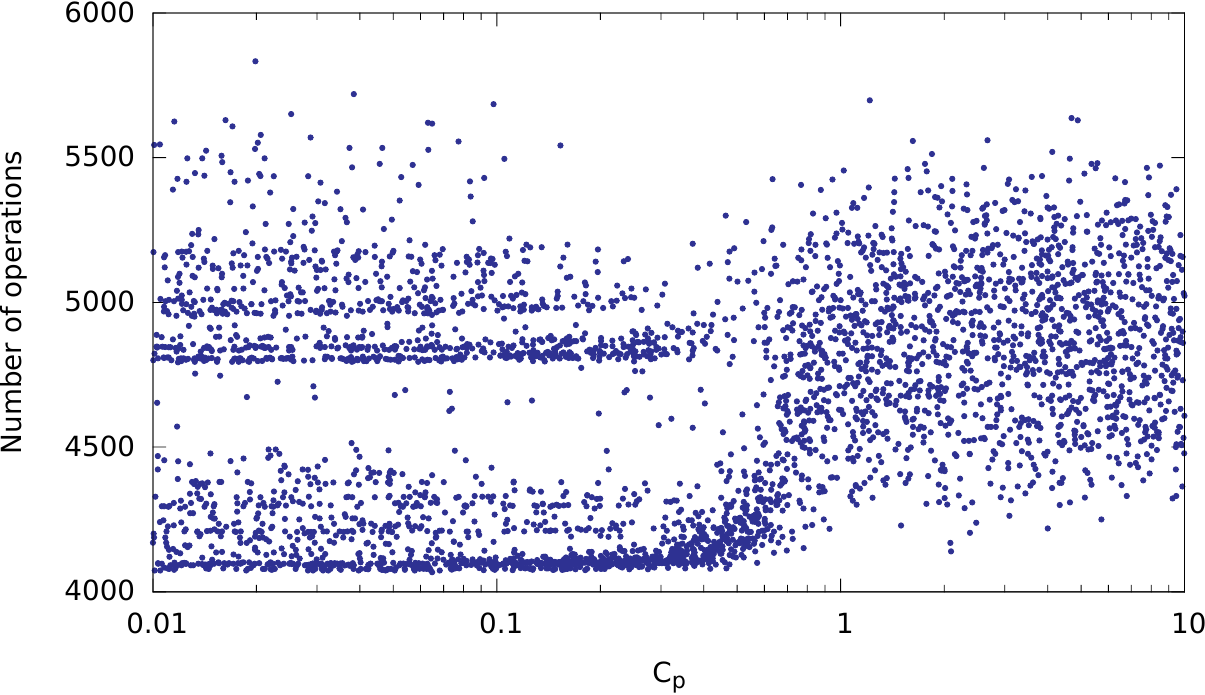}}\\ \bigskip
\subfigure[$N=1000$]{ \includegraphics[scale=0.62]{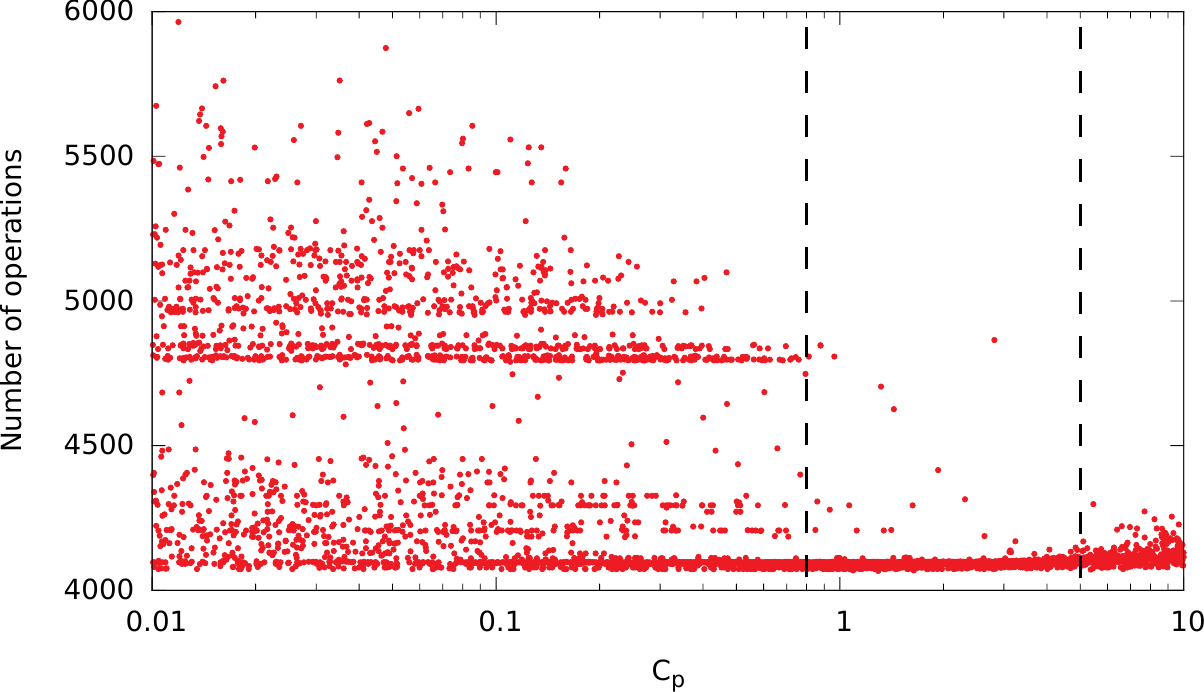}}
\subfigure[$N=1000$]{ \includegraphics[scale=0.62]{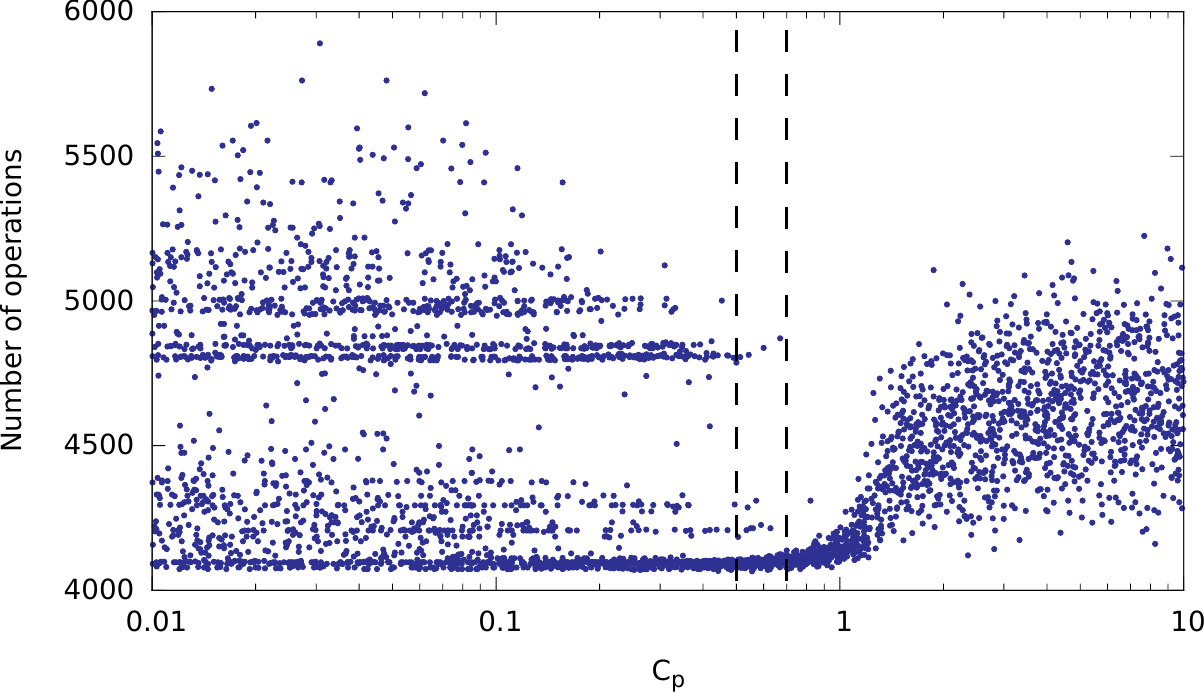}}\\ \bigskip
\subfigure[$N=3000$]{ \includegraphics[scale=0.62]{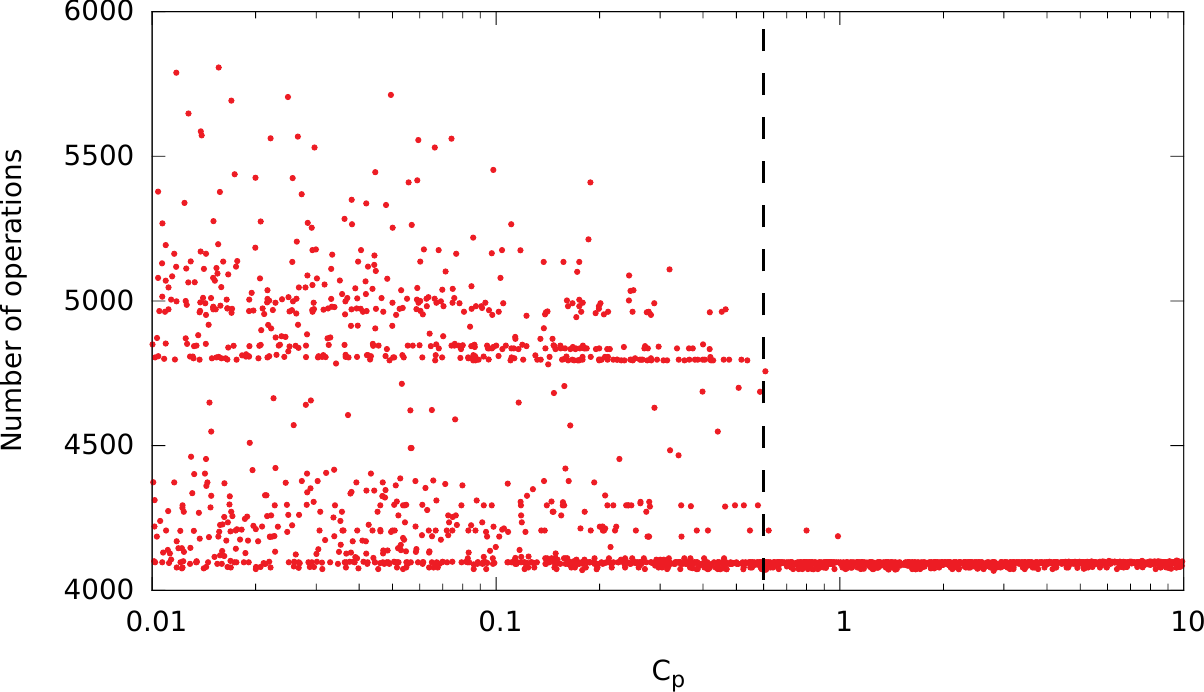}}
\subfigure[$N=3000$]{ \includegraphics[scale=0.62]{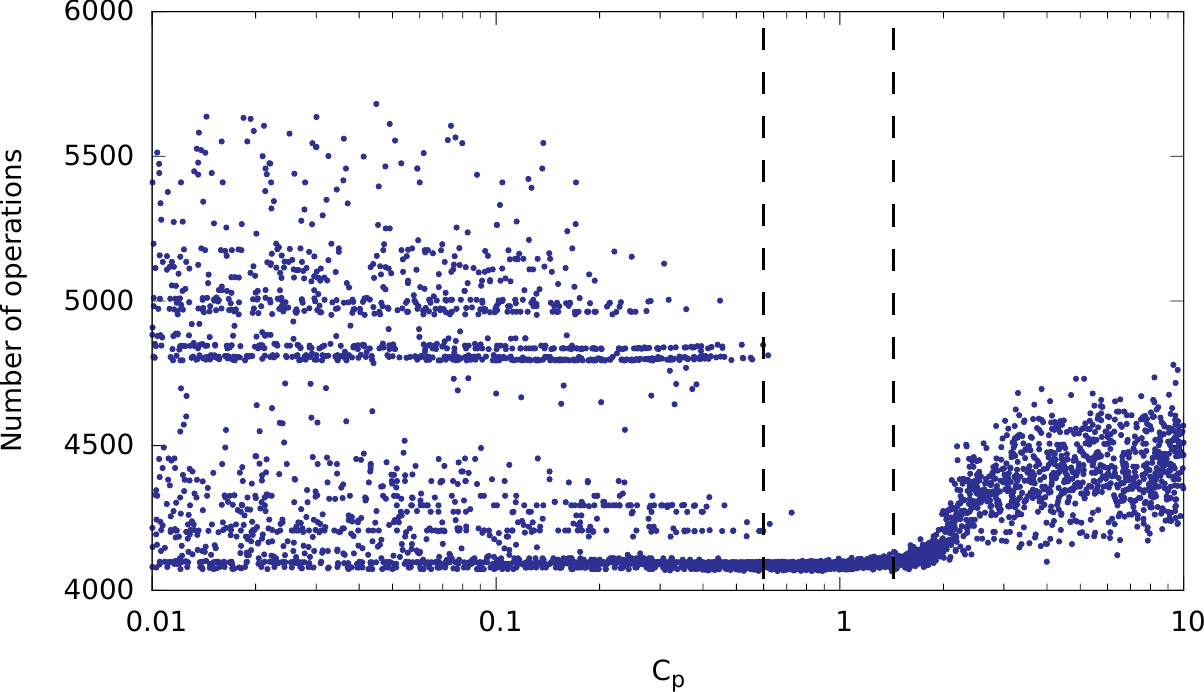}}\\

\caption{HEP($\sigma$) with 15 variables: the number of operations is on the y-axis and $C_p$ on the x-axis. A lower number of operations is better. On the left, we show SA-UCT where $C_p$ is the starting value of $T$ and on the right $C_p$ is the constant in UCT. Each graph contains 4000 runs (dots) of MCTS. Figure \ref{fig:sigma}(a) and \ref{fig:sigma}(b) are measured with $N=300$ tree updates, \ref{fig:sigma}(c) and \ref{fig:sigma}(d) with $N=1000$, and \ref{fig:sigma}(e) and \ref{fig:sigma}(f) with $N=3000$ updates. As indicated by the dashed lines, an area with an operation count close to the global minimum appears, as soon as there are sufficient tree updates $N$. This area is wider for SA-UCT than for UCT.}
\label{fig:sigma}
\end{figure*}

In figure \ref{fig:sigma} we show the results for HEP($\sigma$) with $15$ variables. The figures on the left are generated using SA-UCT (where $C_p$ is the initial temperature). The figures on the right use UCT. Figure \ref{fig:sigma}(a) and \ref{fig:sigma}(b) are measured with $N=300$ tree updates, \ref{fig:sigma}(c) and \ref{fig:sigma}(d) with $N=1000$, and \ref{fig:sigma}(e) and \ref{fig:sigma}(f) with $N=3000$ updates. We see that for both algorithms there are different regions: one region in \ref{fig:sigma}(a) and \ref{fig:sigma}(b), two regions in \ref{fig:sigma}(e) and three regions in \ref{fig:sigma}(c), \ref{fig:sigma}(d), and \ref{fig:sigma}(f). The regions are separated by dashed lines. The regions are called low, intermediate and high. In figure \ref{fig:sigma}(f) these regions are most prominent. At \emph{low} $C_p$ we observe that there are multiple local minima, indicated by high-density band structures (three are prominently visible). At \emph{intermediate} values of $C_p$ we have the region of interest where only the near global minimum is present. At \emph{high} values there is a diffuse region with no distinguishable local minima. This happens when there is too much exploration.

Comparing the graphs on the left and on the right, we see that the linearly decreasing $C_p$ causes a horizontal stretching, which makes the region of interest larger. If we look at the middle graphs, \ref{fig:sigma}(c) and \ref{fig:sigma}(d), where the number of updates $N=1000$, the region of interest is approximately $[0.8,5.0]$ for a linearly decreasing $C_p$, whereas it is roughly $[0.5,0.7]$ for a constant $C_p$. Thus, SA-UCT makes the region of interest about $20$ times larger for HEP($\sigma$), relative to the uninteresting low $C_p$ region with local minima which did not grow significantly. For $N=3000$, the difference in size of the region of interest is even larger.

\begin{figure*}
\centering
\textbf{res(7,5) with 14 variables} \par\medskip
\subfigure[$N=300$]{ \includegraphics[scale=0.62]{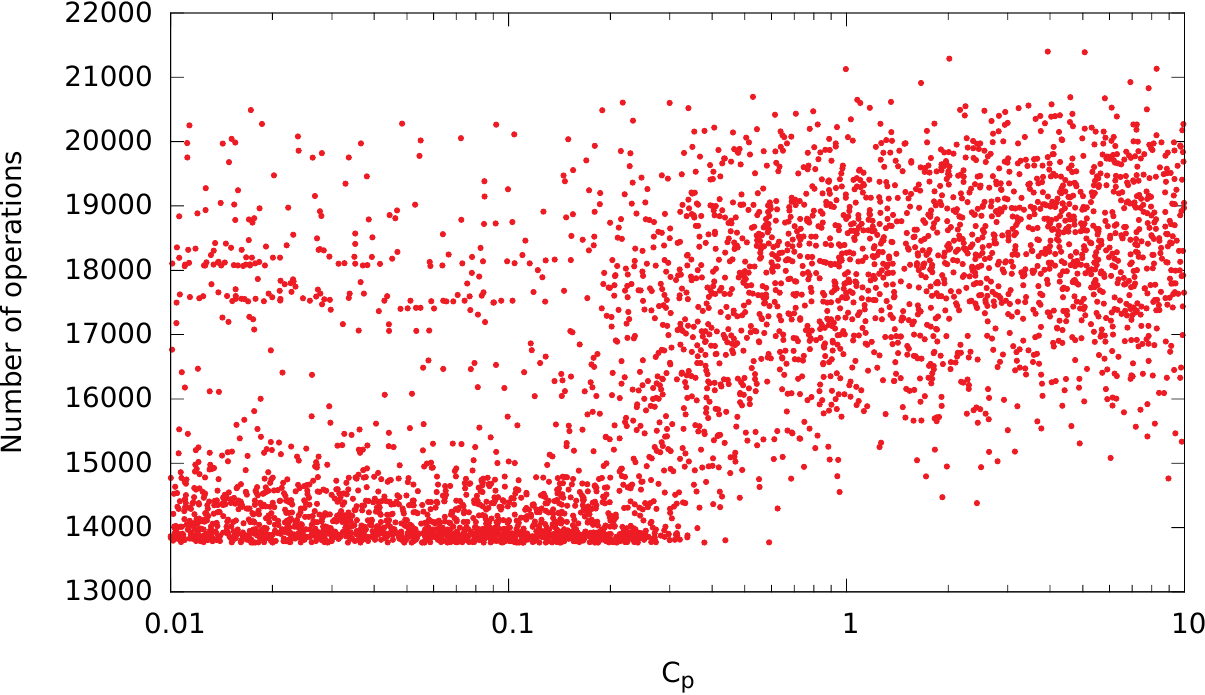} }
\subfigure[$N=300$]{ \includegraphics[scale=0.62]{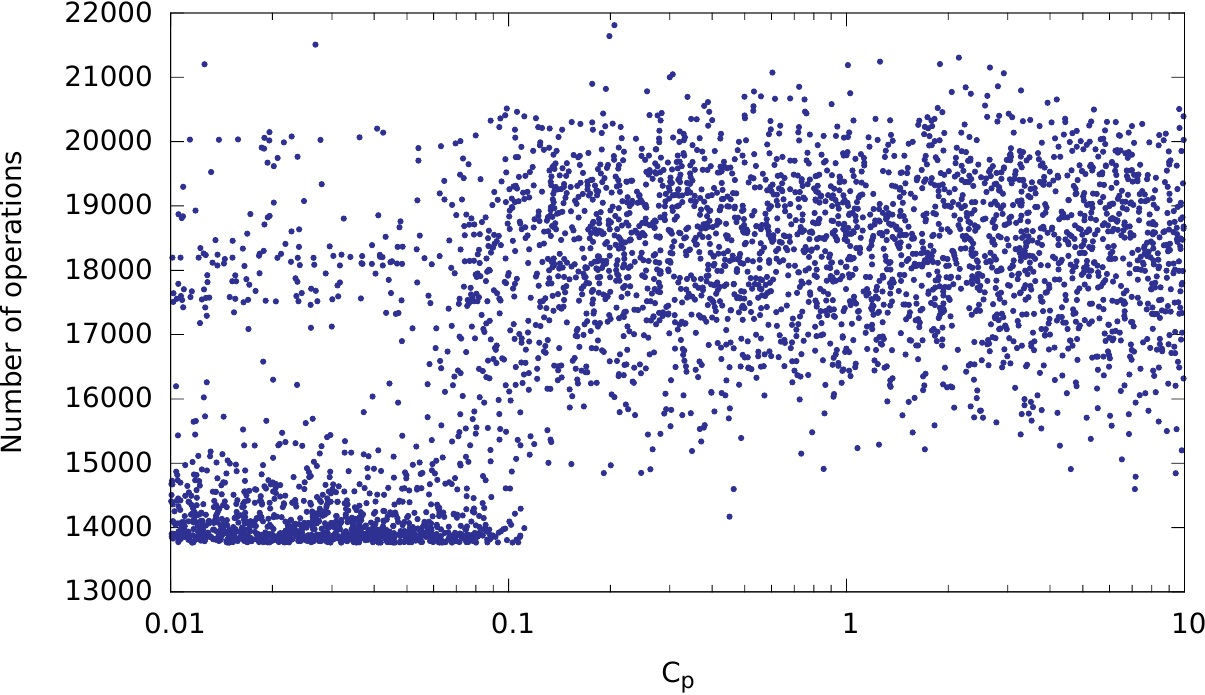} }\\ \bigskip
\subfigure[$N=1000$]{ \includegraphics[scale=0.62]{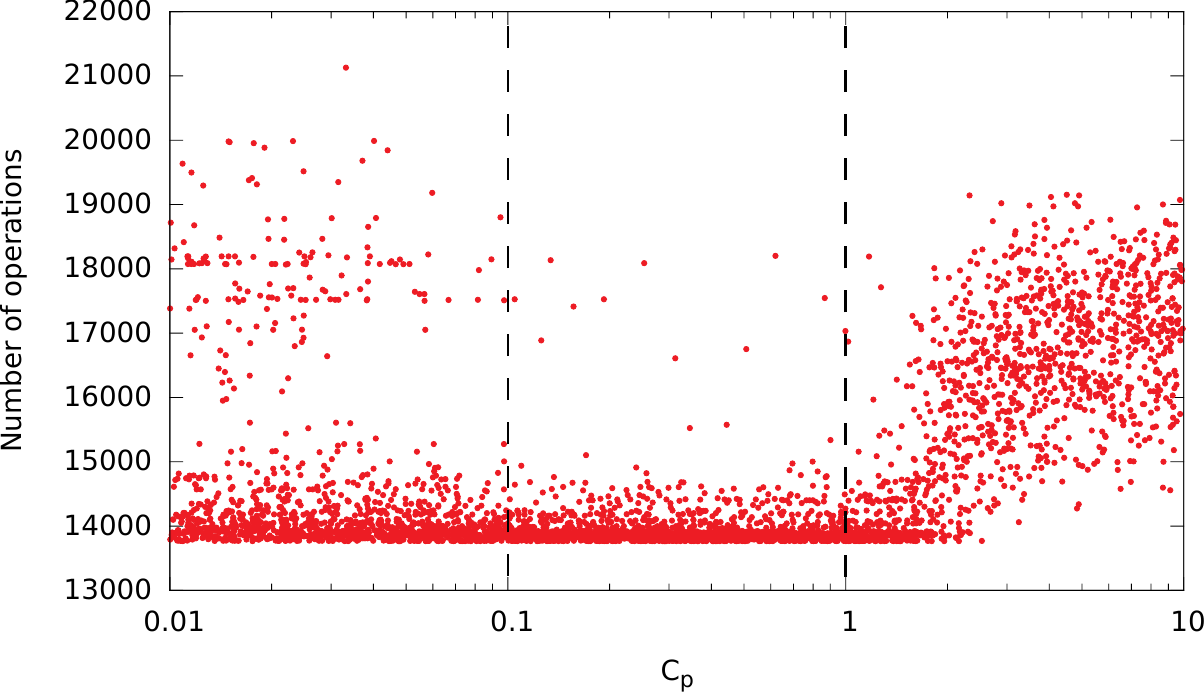} }
\subfigure[$N=1000$]{ \includegraphics[scale=0.62]{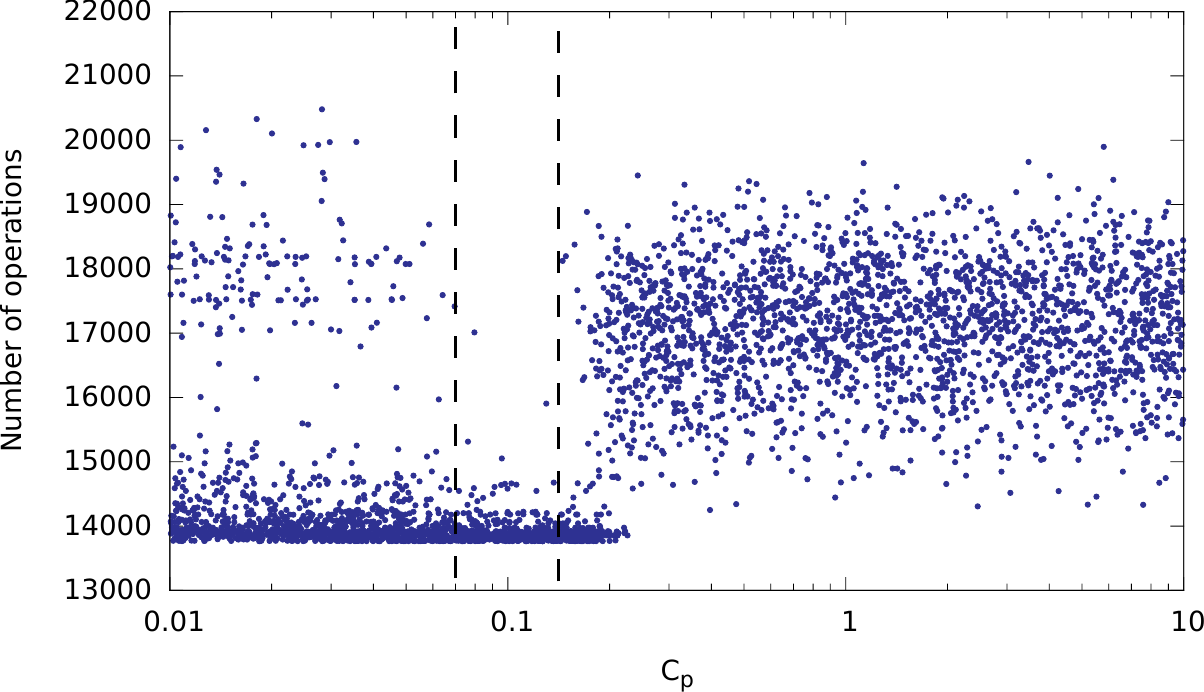} }\\ \bigskip
\subfigure[$N=3000$]{ \includegraphics[scale=0.62]{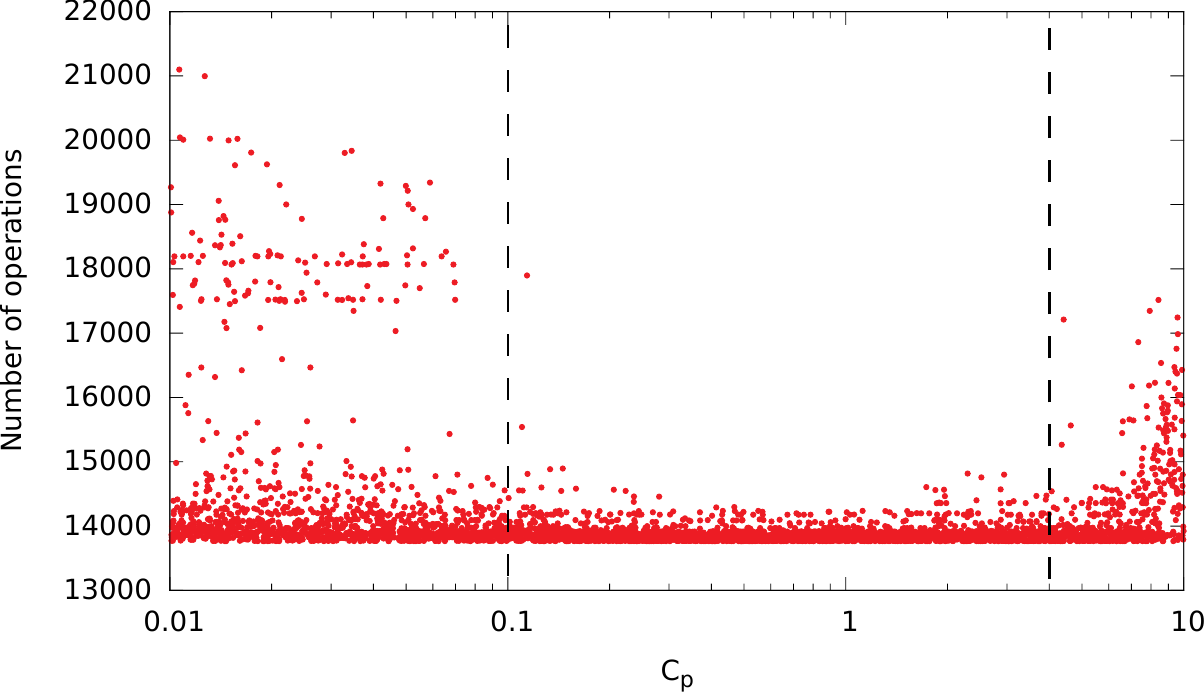} }
\subfigure[$N=3000$]{ \includegraphics[scale=0.62]{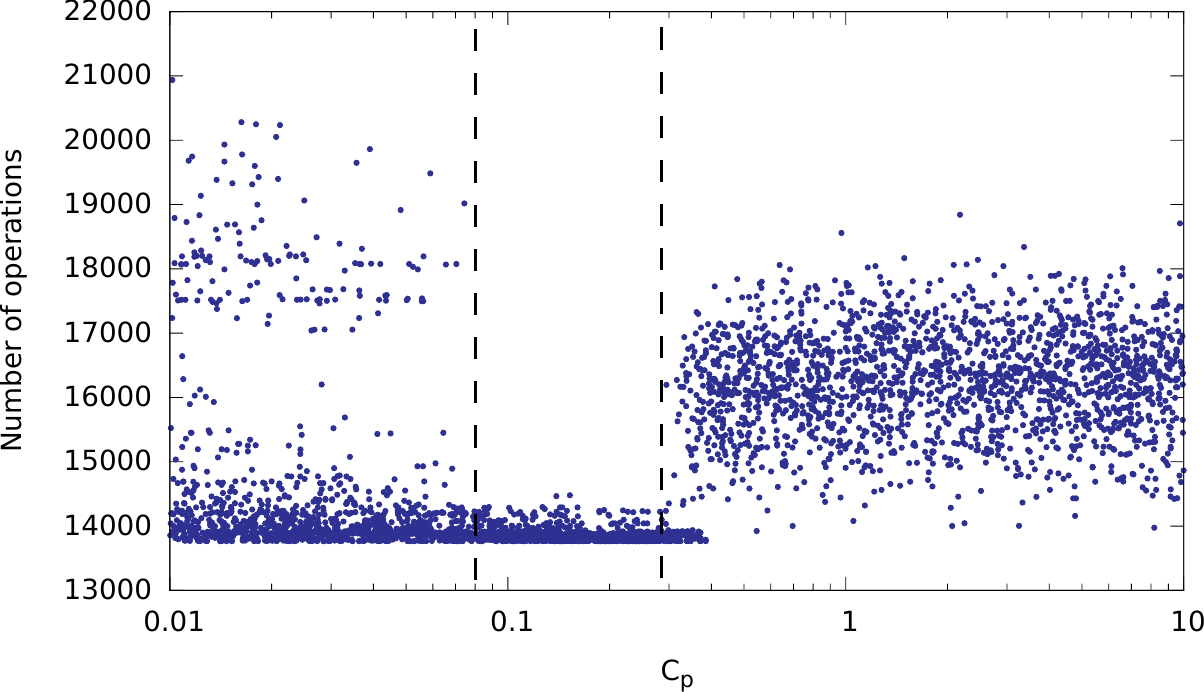} }\\
\caption{res(7,5) polynomial with 14 variables: the number of operations is on the y-axis and $C_p$ on the x-axis. A lower number of operations is better. On the left, we show SA-UCT where $C_p$ is the starting value of $T$ and on the right $C_p$ is the constant in UCT. Each graph contains 4000 runs (dots) of MCTS. Figure \ref{fig:res75}(a) and \ref{fig:res75}(b) are measured with $N=300$ tree updates, \ref{fig:res75}(c) and \ref{fig:res75}(d) with $N=1000$, and \ref{fig:res75}(e) and \ref{fig:res75}(f) with $N=3000$ updates. As indicated by the dashed lines, an area with an operation count close to the global minimum appears, as soon as there are sufficient tree updates $N$. This area is wider for SA-UCT than for UCT.}
\label{fig:res75}
\end{figure*}

\begin{figure*}
\centering
\textbf{F13 with 22 variables} \par\medskip
\subfigure[$N=300$]{ \includegraphics[scale=0.62]{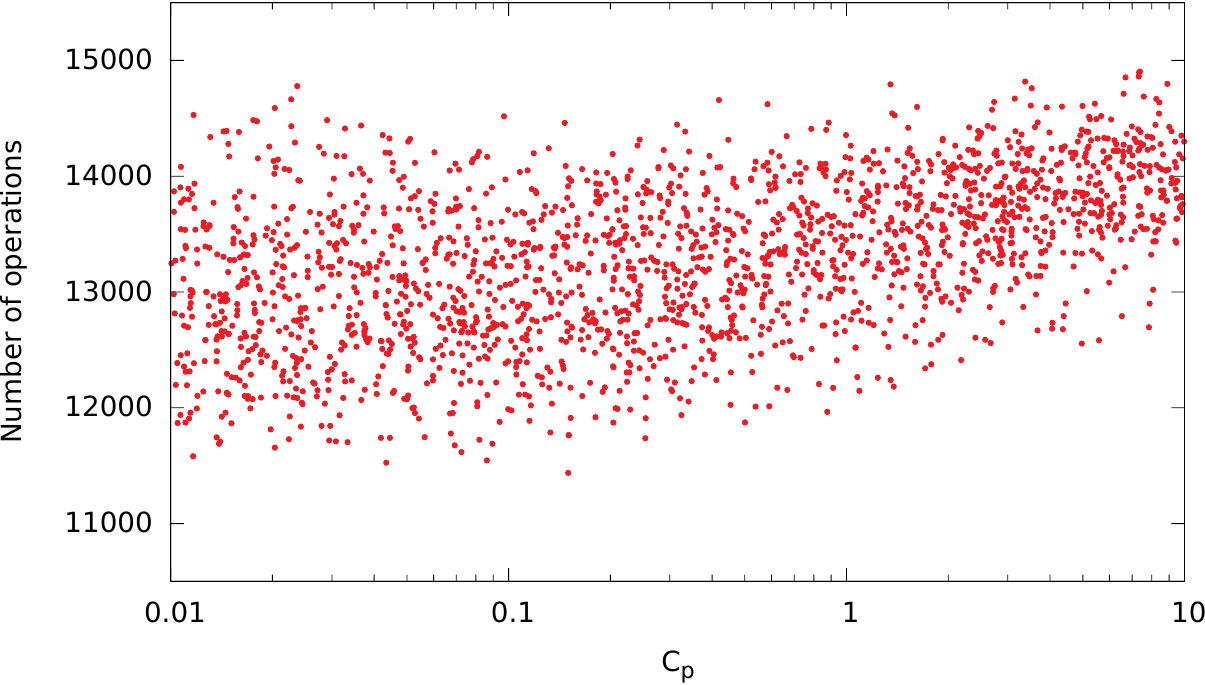} }
\subfigure[$N=300$]{ \includegraphics[scale=0.62]{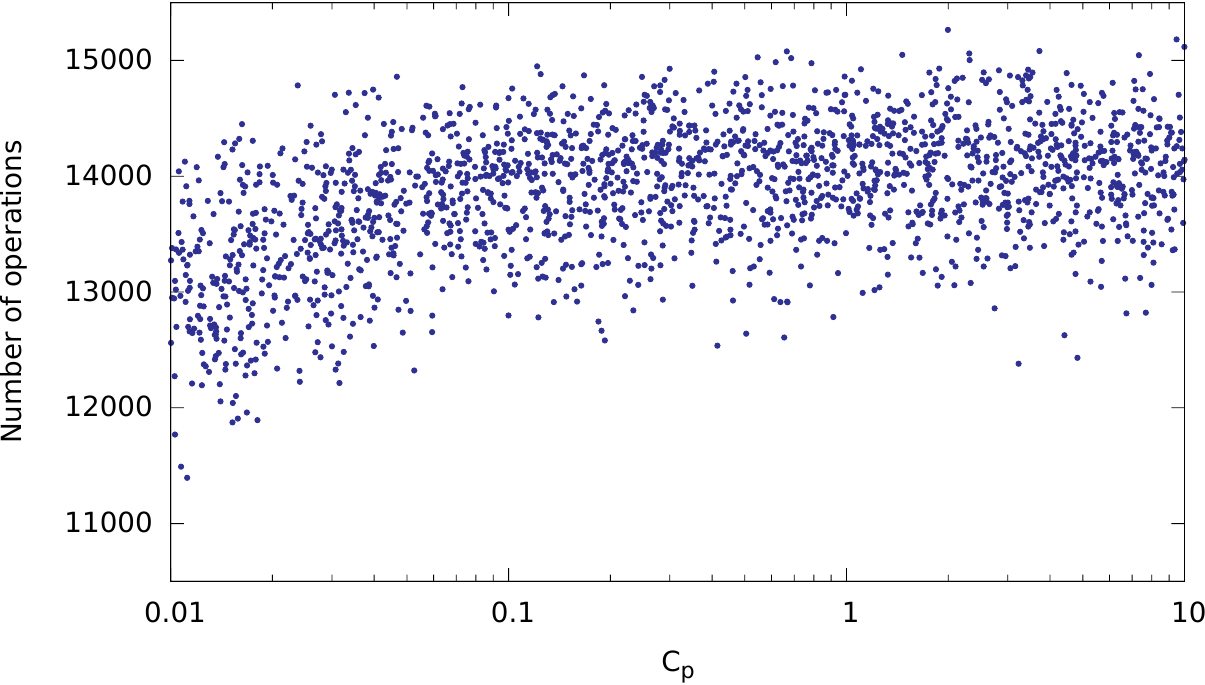} }\\ \bigskip
\subfigure[$N=1000$]{ \includegraphics[scale=0.62]{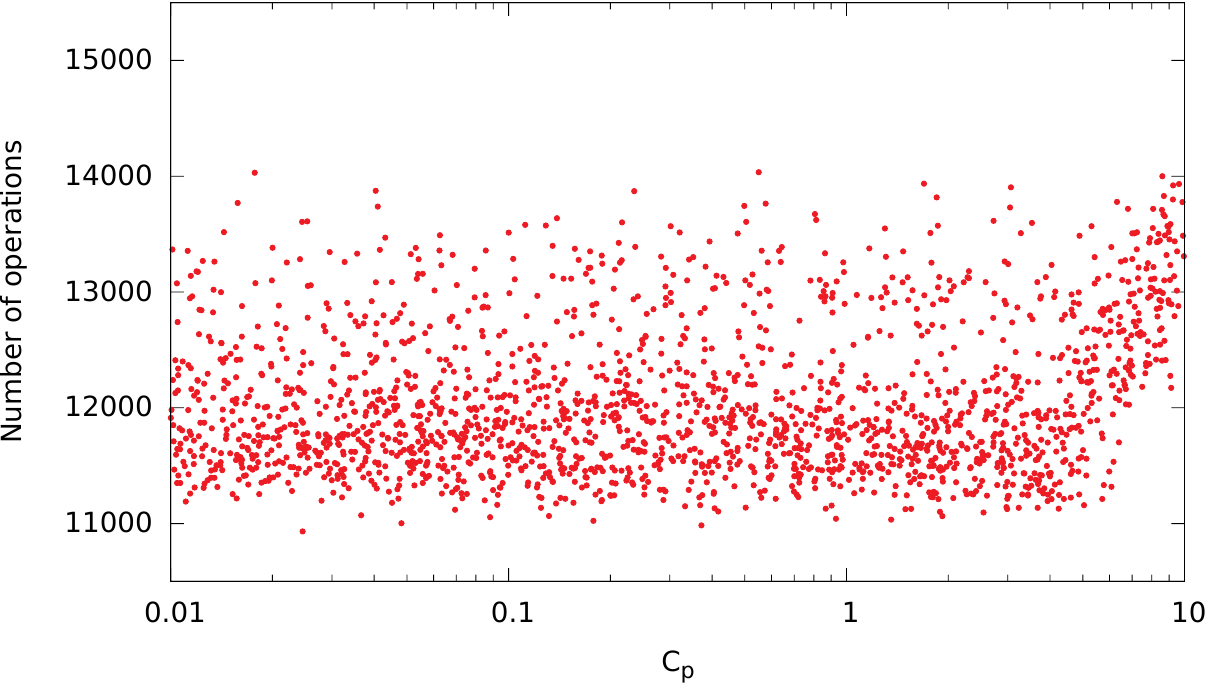} }
\subfigure[$N=1000$]{ \includegraphics[scale=0.62]{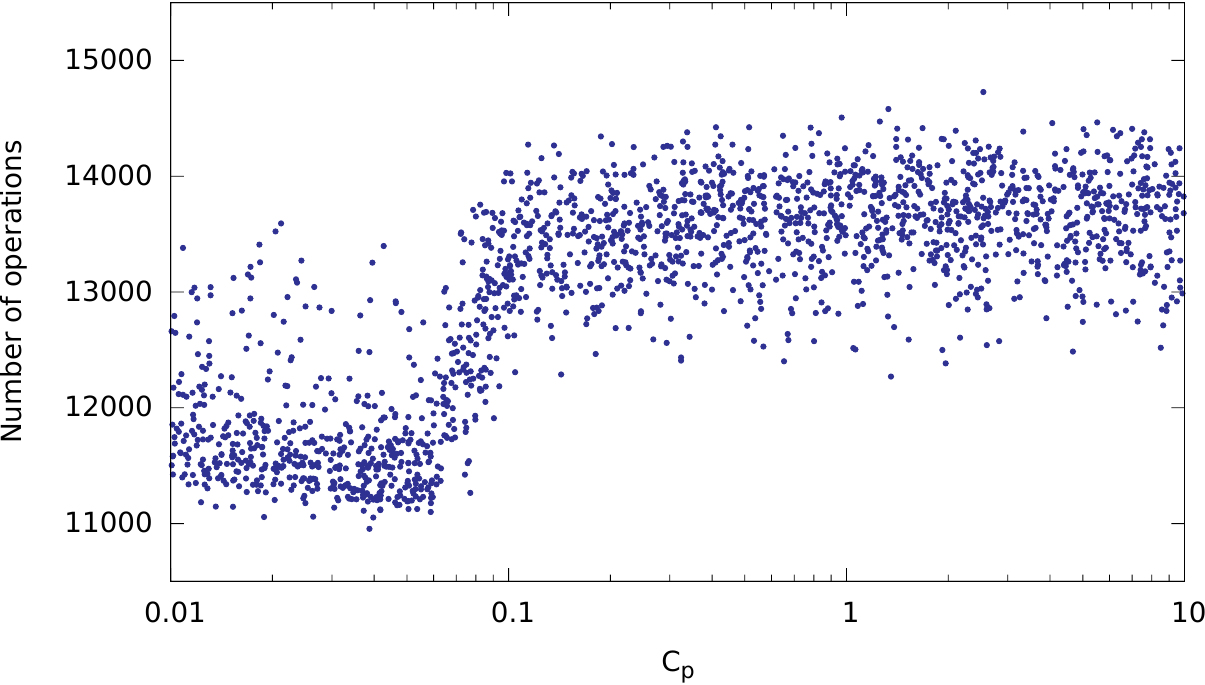} }\\ \bigskip
\subfigure[$N=3000$]{ \includegraphics[scale=0.62]{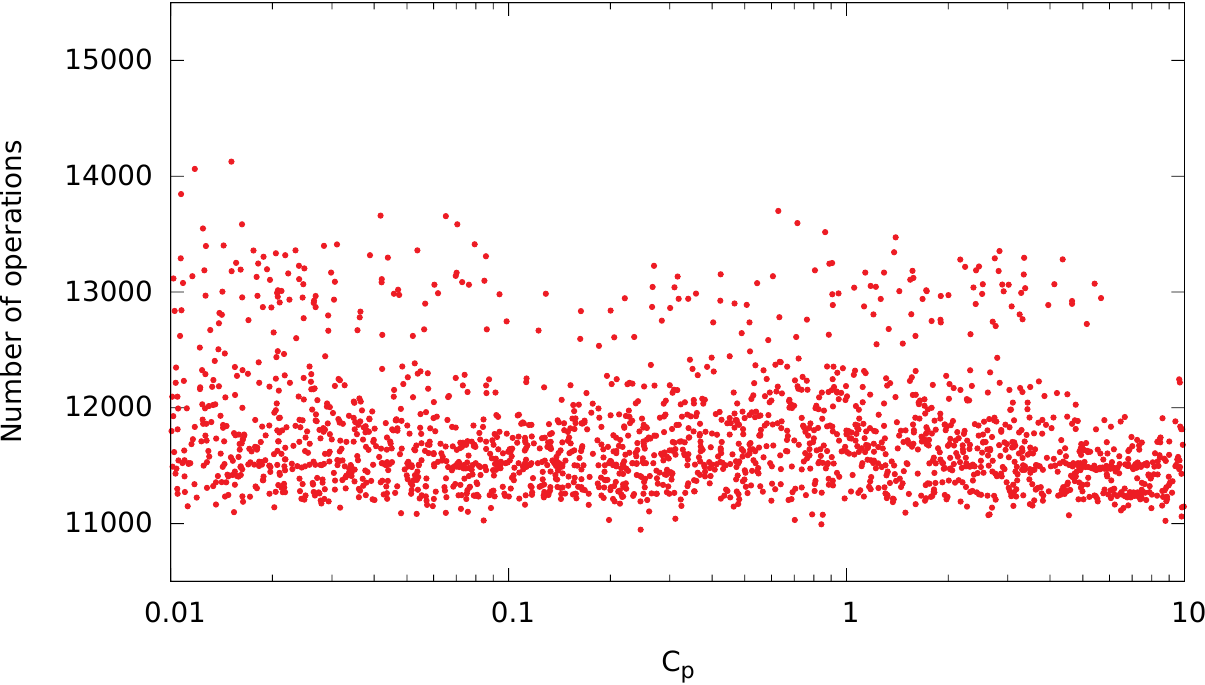} }
\subfigure[$N=3000$]{ \includegraphics[scale=0.62]{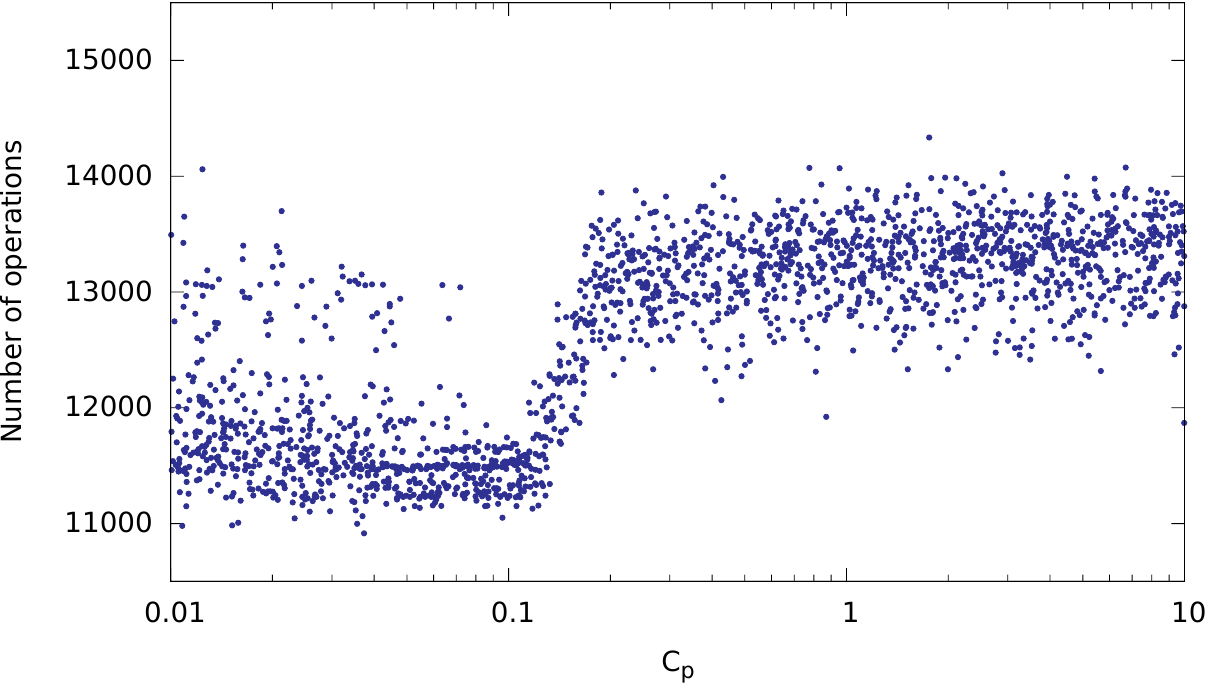} }\\
\caption{F13 with 22 variables: the number of operations is on the y-axis and $C_p$ on the x-axis. A lower number of operations is better. On the left, we show SA-UCT where $C_p$ is the starting value of $T$ and on the right $C_p$ is the constant in UCT. Each graph contains 4000 runs (dots) of MCTS. Figure \ref{fig:sigma}(a) and \ref{fig:sigma}(b) are measured with $N=300$ tree updates, \ref{fig:sigma}(c) and \ref{fig:sigma}(d) with $N=1000$, and \ref{fig:sigma}(e) and \ref{fig:sigma}(f) with $N=3000$ updates. F13 exhibits slightly different behavior from HEP($\sigma$) in figure \ref{fig:sigma}. There are no band structures at low $C_p$, but there are two diffuse regions: one near the global minimum at low $C_p$ and one at a higher local minimum. Using SA-UCT, the region near the global minimum is wider than for UCT.}
\label{fig:f13}
\end{figure*}

In figure \ref{fig:res75} we have tested our method on an expression from the field of mathematics, namely a resultant res(7,5), where $\text{res}(m, n) = \text{res}_x(\sum^m_{i=0} a_ix^i, \sum^n_{i=0} b_ix^i)$, as described in \cite{Leiserson2010}. While from a different field, we still observe the band structures at low $C_p$ and the widening of the region of interest occurs here as well. For figure \ref{fig:res75}(c) with $N=1000$ tree updates using SA-UCT the region of interest is approximately $[0.1,1.0]$ and for \ref{fig:res75}(d) using UCT it is approximately $[0.07,0.15]$. This means that the region of interest has become about 10 times wider. 
In section \ref{sec:futurework} we continue our findings on this polynomial in a discussion.

In figure \ref{fig:f13} we show the results for our third expression, called F13, which again stems from the field of high energy physics and has 22 variables. Since the depth of a complete tree is equal to the number of variables, more tree updates are required to reach the final node for F13, compared to the other two expressions. From the graphs we see that structure emerges around $N=1000$. For UCT, there are two clouds: one near the global minimum and one near a higher local minimum. Contrary to the previous two expressions, F13 does not have a band structure at low $C_p$, but exhibits a diffuse cloud near the global minimum. However, we see that this cloud is wider (roughly 50 times at $N=1000$) for SA-UCT than for UCT, as was the case for HEP($\sigma$) and res(7,5). Since the band is still broad, multiple samples are required to approach the global minimum, regardless of $C_p$. This is governed by the $R$ parameter \cite{Herik2013B,Kuipers2013B}, and is consequently a topic for future research. 

\section{\uppercase{Conclusion}}
\label{sec:conclusion}

\noindent In this work we proposed a new UCT formula, called SA-UCT, that has a decreasing exploration-exploitation parameter $T$, similar to the temperature in simulated annealing. We have compared the performance of SA-UCT to the performance of UCT using three large expressions from physics and mathematics. From our experimental results we may provisionally conclude that SA-UCT significantly increases the range of initial temperatures $C_p$ for which good results are obtained. This facilitates the selection of an appropriate $C_p$.

During our research, we uncovered multiple areas for future research.

\section{\uppercase{Discussion / future work}}
\label{sec:futurework}

\begin{figure}[ht]
\centering
\subfigure[Forward scheme]{ \includegraphics[scale=0.6]{Data/res75_1000_fwd.pdf} \label{fig:backa}}\\ \smallskip
\subfigure[Backward scheme]{ \includegraphics[scale=0.6]{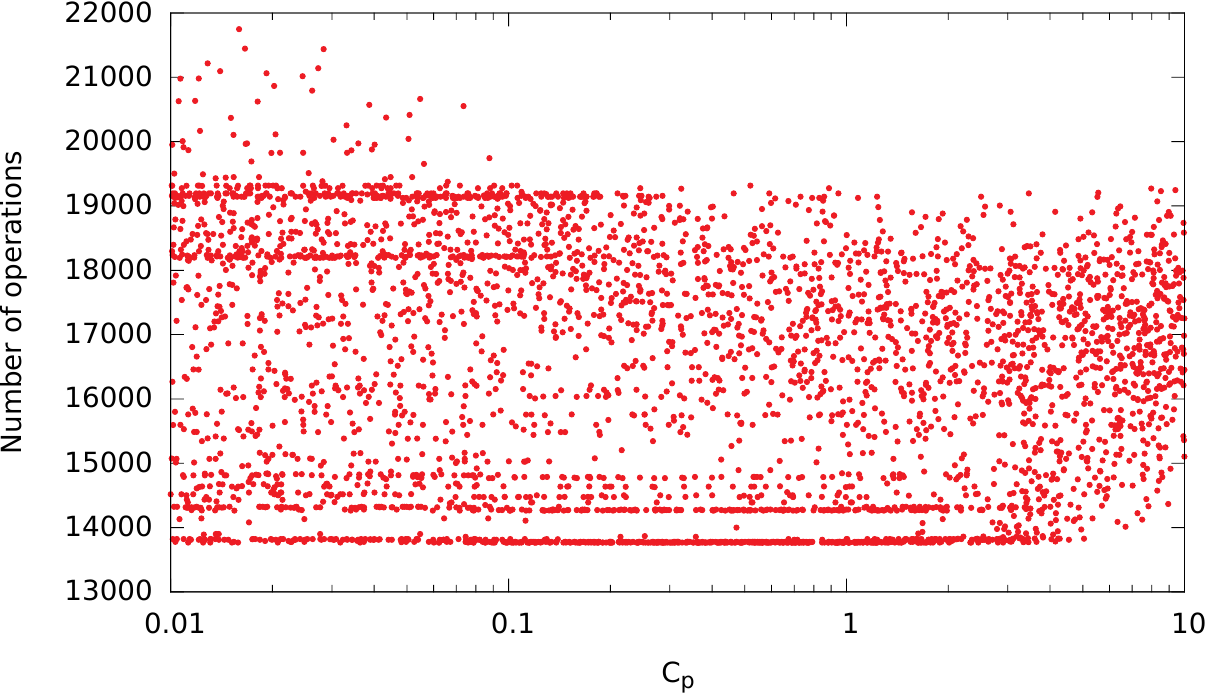} \label{fig:backb}}
\caption{res(7,5): differences between forward (top) and backward (bottom) Horner schemes, at $N=1000$ tree updates with SA-UCT. Forward Horner schemes generate a region of $C_p$ where the number of operations is near the global minimum, whereas backward schemes have multiple high-density local minima and a diffuse region.}
\label{fig:fwd_vs_bkwd}
\end{figure}

\noindent We start the discussion by distinguishing normal Horner scheme constructions from reversed constructions, also called forward and backward respectively. In the backward scheme, we create the Horner scheme from the inside out, reversing the extraction order. For example, in eq. \eqref{eq:horner} the forward scheme is $y, x$ and the backward scheme is $x, y$. The distinction between the two constructions is important for MCTS, because by its nature the tree of MCTS is asymmetric: the children of the root are all explored, but most nodes at the bottom will not. Since we are interested in the entire path, this means that the end of the path will be underexplored compared to the beginning of the path. If large improvements can be made by carefully selecting variables at the end of the scheme, these optimizations will likely not be found. Figure \ref{fig:fwd_vs_bkwd} illustrates the effect that a forward and a backward scheme have on the res(7,5) expression, where $N=1000$ and SA-UCT is used. A forward scheme yields the three regions mentioned in section \ref{sec:results}, whereas the backward scheme yields multiple local minima and a diffuse area for every $C_p$.
The difference between forward and backward schemes is present for both SA-UCT and UCT, although it is more prominent in the latter. For UCT with $C_p > 0.1$, the tree often does not reach the end if the number of variables is larger than 15. The path is then completed using the random default policy, which selects a single path and consequently does no exploration. For SA-UCT, the tree often does reach the end and some exploration occurs, because a low and exploitative $C_p$ effectively explores deeper in the tree. However, this effect is not sufficient to smooth out the differences between forward and backward schemes, as can be seen in figure \ref{fig:fwd_vs_bkwd}.
We found that choosing a backward scheme for HEP($\sigma$) and F13 leads to significant improvements. Whether we can predict beforehand (by looking at the expression) if forward or backward search has to be used is a topic of current research. 
Additionally, other ways than forward or backward construction of the Horner schemes could be used. For example, one could put more emphasis on the middle part of the scheme, by making the first variables map to the middle of the Horner scheme and working outwards from the center. Here again additional work is needed to predict which Horner scheme construction works best.

Also, more research is needed to find quickly a value of $C_p$ in the region of interest. If the number of tree updates $N$ is sufficiently high, the region of interest becomes so large that even a binary search may be sufficient to find a good $C_p$. In order to understand what $N$ is required to obtain such a large region of interest, the relation between an adequate $N$ and the number of variables has to be further investigated.

Furthermore, the performance of SA-UCT has to be measured for different applications. Examples are the travelling salesman problem and Go. Many Go implementations currently set $C_p=0$, effectively disabling UCT, but perhaps a small value for $C_p$ is fruitful if SA-UCT is applied \cite{Lee2009}.

Moreover, additional work is needed to examine different schemes for decreasing $C_p$. For example, the current depth in the tree may be a good candidate. One other possibility is detecting if the best child selection gets stuck in a local minimum and `over-explores' a branch. If this is the case, the $T$ could be increased to find further minima in unexplored branches. Different cooling functions could also be tried, such as exponentially decreasing $T$.

Finally, we believe that the use of domain specific knowledge can be fruitfully explored if the expression has sufficient structure. To confirm this belief more research is needed.


\bibliographystyle{apalike}
{\small
\bibliography{ref}}

\end{document}